\documentclass[sigconf]{acmart}
\AtBeginDocument{%
  }

\copyrightyear{2023} 
\acmYear{2023} 
\setcopyright{acmlicensed}\acmConference[CIKM '23]{Proceedings of the 32nd
ACM International Conference on Information and Knowledge
Management}{October 21--25, 2023}{Birmingham, United Kingdom}
\acmBooktitle{Proceedings of the 32nd ACM International Conference on
Information and Knowledge Management (CIKM '23), October 21--25, 2023,
Birmingham, United Kingdom}
\acmPrice{15.00}
\acmISBN{979-8-4007-0124-5/23/10}
\acmDOI{XXXXXXX.XXXXXXX}

\usepackage{algorithm}
\usepackage{url}
\usepackage{balance}
\usepackage{algorithmic}

\begin{document}


\title[Temporal Convolutional Explorer]{Temporal Convolutional Explorer Helps Understand 1D-CNN's Learning Behavior in Time Series Classification from Frequency Domain}

\author{Junru Zhang}
\affiliation{%
  \institution{Zhejiang University}
  \city{Hangzhou}
  \country{China}
}
\email{junruzhang@zju.edu.cn}

\author{Lang Feng}
\affiliation{%
  \institution{Zhejiang University}
  \city{Hangzhou}
  \country{China}
}
\email{langfeng@zju.edu.cn}

\author{Yang He}
\affiliation{%
  \institution{Zhejiang University}
  \city{Hangzhou}
  \country{China}
}
\email{he_yang@zju.edu.cn}

\author{Yuhan Wu}
\affiliation{%
  \institution{Zhejiang University}
  \city{Hangzhou}
  \country{China}
}
\email{wuyuhan@zju.edu.cn}

\author{Yabo Dong}
\affiliation{%
  \institution{Zhejiang University}
  \city{Hangzhou}
  \country{China}
}
\authornote{Corresponding author}
\email{dongyb@zju.edu.cn}

\renewcommand{\shortauthors}{Junru Zhang, Lang Feng, Yang He, Yuhan Wu, \& Yabo Dong}
\newtheorem{remark}{Remark}
\begin{abstract}
  While one-dimensional convolutional neural networks (1D-CNNs) have been empirically proven effective in time series classification tasks, we find that there remain undesirable outcomes that could arise in their application, motivating us to further investigate and understand their underlying mechanisms. In this work, we propose a Temporal Convolutional Explorer (TCE) to empirically explore the learning behavior of 1D-CNNs from the perspective of the frequency domain. Our TCE analysis highlights that deeper 1D-CNNs tend to distract the focus from the low-frequency components leading to the accuracy degradation phenomenon, and the disturbing convolution is the driving factor. Then, we leverage our findings to the practical application and propose a regulatory framework, which can easily be integrated into existing 1D-CNNs. It aims to rectify the suboptimal learning behavior by enabling the network to selectively bypass the specified disturbing convolutions. Finally, through comprehensive experiments on widely-used UCR, UEA, and UCI benchmarks, we demonstrate that 1) TCE's insight into 1D-CNN's learning behavior; 2) our regulatory framework enables state-of-the-art 1D-CNNs to get improved performances with less consumption of memory and computational overhead.
\end{abstract}

\ccsdesc[500]{Information systems~Data mining}
\ccsdesc[500]{Computing methodologies~Neural networks}

\keywords{time series classification, one-dimensional convolutional neural networks, accuracy degradation, learning behavior}


\maketitle
 
\section{Introduction}
\begin{figure}[t]
  \centering
  \includegraphics[width=0.88\columnwidth,height=0.454\linewidth]{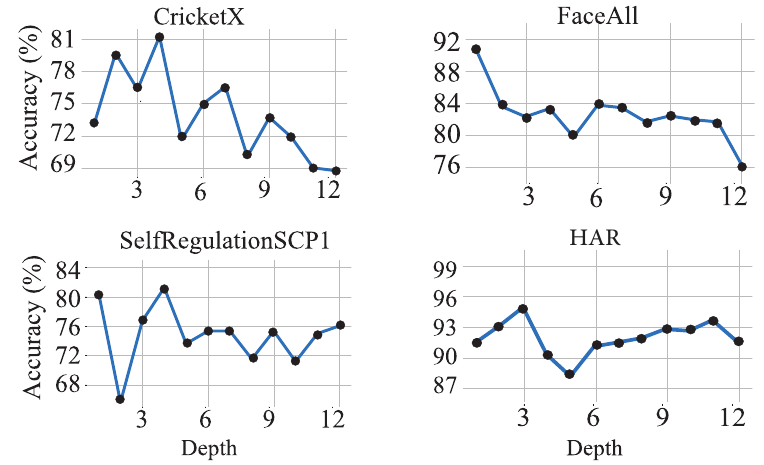} 
  \caption{Accuracy of ResNet at different depths on four TSC datasets. The highest accuracies of all datasets do not occur at the maximum depth.}
  \label{basic_fig}
\end{figure}
In the past, time series classification (TSC) tasks were commonly addressed by traditional methods, such as distance-based \cite{shokoohi2017generalizing}, similarity-based \cite{bagnall2017great}, interval-based \cite{fulcher2017hctsa} and shapelet-based techniques \cite{grabocka2014learning}. Unfortunately, these methods require tedious crafting on feature engineering or data preprocessing. With the empirical successes of deep learning in the computer vision (CV) field, some researchers began to explore the application of deep neural networks (DNNs) in TSC, especially convolutional neural networks (CNNs) to 1D time series domain, such as Fully Convolutional Network (FCN) \cite{wang2017time}, Residual Networks (ResNet) \cite{wang2017time,ruiz2021great}, and InceptionTime \cite{ismail2020inceptiontime}. CNN-based methods offer the advantage of avoiding the need for elaborate feature engineering and data preprocessing and are generally more flexible to be applied to various time series scenarios. However, unlike CNNs handling 2D image data which has been heavily studied, the working mechanism of one-dimensional CNNs (1D-CNNs) on TSC has received relatively little attention, indicating that they have the remaining power to be inspired \cite{zerveas2021transformer}. 

In the CV field, it is a well-established empirical trend that deeper CNNs tend to exhibit stronger performance \cite{he2016deep,2016Rethinking,arora2018optimization}. However, upon revisiting this property of 1D-CNNs, we find that they do not follow this trend in TSC tasks. To shed light on this intriguing phenomenon, we designed an intuitive experiment examining the relationship between accuracy and network depth. Specifically, we employed the ResNet architecture as a foundational network to mitigate issues such as network degradation and gradient vanishing \cite{he2016deep}. As indicated in Fig. \ref{basic_fig}, contrary to a common view, we unexpectedly found that \emph{deeper $\ne$ better}. 
Furthermore, the analysis, as discussed in Sec. \ref{secoverfitting}, has revealed that overfitting does not explain the observed phenomenon, thereby implicating the involvement of other factors.    
Such noteworthy outcomes pertaining to the accuracy degradation issue have motivated our investigation into the intricate learning mechanisms of deep 1D-CNNs. Our goal is to develop a comprehensive understanding of their learning behavior and identify the underlying factors at play. Through this exploration, effective strategies can be devised to improve the performance of deep 1D-CNNs for TSC tasks.

Starting from some usual observations in deep CNNs, recent studies have delved into the underlying reasons behind the training outcomes of deep CNNs, focusing on analyzing the correlation between the frequency spectrum of images and deep CNNs' behavior \cite{xu2019training,rahaman2019spectral,xu2021deep,wang2020high}.
Although incorporating frequency analysis methods into time series data with a wealth of frequency characteristics analogous to images \cite{schafer2017multivariate,yang2022unsupervised}, could aid in elucidating the causes of the aforementioned phenomenon, such approaches are not inherently applicable for 1D-CNNs in TSC tasks. 
The primary distinction lies in the fact that 1D-CNNs process the time series data relying on capturing temporal information between numerous channels. Specifically, as opposed to image data that only contains three color channels, TSC tasks typically have a high number of channels with diverse attributes \cite{bai2021correlative}. 
1D-CNNs focus on extracting temporal features by recognizing the cross-channel information in time series, while CNNs focus on capturing spatial features by processing inter-pixel relationships within images. These indicate the learning behavior of 1D-CNNs for TSC tasks is distinct, thereby highlighting the necessity of a novel mechanism to explore it.

To this end, we provide the \emph{Temporal Convolutional Explorer (TCE)} mechanism to identify the frequency components of time series that are emphasized/overlooked by deeper convolutional layers. By conducting Fast Fourier Transform (FFT) \cite{1965An} on each channel of the feature map and the input instance, TCE reveals that deeper 1D-CNNs \emph{distract the focus from low frequencies} leading to the accuracy degradation, and points out that \emph{the disturbing convolution} is the driving factor for this problem. To leverage our findings to a practical application, we further propose a plug-and-play \emph{regulatory framework} composed of TCE and gating mechanism to rectify the suboptimal learning behavior by selectively skipping over the specified disturbing convolutions. We verify TCE's insights through a series of comprehensive experiments on public UCR \cite{dau2019ucr}, UEA \cite{bagnall2018uea} and UCI \cite{asuncion2007uci} benchmarks, including univariate TSC (UTSC) datasets and multivariate TSC (MTSC) datasets. The experimental results first demonstrate that deeper 1D-CNNs exhibit stronger \emph{learning ability} (i.e., generalization performance and learning speed) on low frequencies. To understand why the deeper network with such strong learning ability does not perform well, we then experimentally reveal that deeper 1D-CNNs tend to distract the focus from the low frequencies. Such \emph{learning behavior} could hinder the deep network from fully utilizing its learning ability, ultimately resulting in accuracy degradation. After skipping the disturbing convolutions, deeper 1D-CNNs effectively improve the accuracy and recover the focus on low frequencies, which supports that the disturbing convolution is responsible for low-frequency focus distraction. To further verify our practical application, we equip our regulatory framework on advanced 1D-CNN baselines (i.e., ResNet, InceptionTime, and FCN), and the results show that our framework enables these models to improve their performances with less consumption of memory and computational overhead. It is important to note that the aim of our work is \emph{not} to claim a new algorithm for TSC. Instead, we attempt to provide empirical analysis to help understand the learning behavior of 1D-CNNs on TSC tasks and present a practical application of our findings. We hope TCE's insights can benefit the community to develop more suitable and powerful 1D-CNN classifiers for TSC tasks.

\section{Related Work }
Deep learning methods, particularly 1D-CNNs, have been extensively explored in the analysis of 1D time series. \cite{wang2017time} first made a comparison of the FCN and ResNet on 44 UTSC tasks, and then \cite{2019Deep} provided the standardized large-scale comparative study of deep learning approaches in TSC. They found that FCN and ResNet could significantly be better than all other methods on UTSC and MTSC datasets. Building upon the concept of Inception modules \cite{szegedy2015going}, InceptionTime was introduced as an advanced CNN-based approach for UTSC tasks \cite{ismail2020inceptiontime}. The effectiveness of InceptionTime was further confirmed by \cite{ruiz2021great} in their comparison of MTSC algorithms on the UEA archive, where InceptionTime outperformed traditional methods like Dynamic Time Warping \cite{shokoohi2017generalizing}.

In addition, some effective methods \cite{dempster2020rocket,dempster2021minirocket,tan2022multirocket} aim to use the random 1D convolutional kernels without training, thus being considered as non-deep learning methods \cite{ruiz2021great,zerveas2021transformer}. Nevertheless, their success emphasizes the potential of 1D convolutional kernels to filter time series features, especially frequency information \cite{pantiskas2022taking}. In addition, some methods \cite{schafer2012sfa,schafer2015boss,schafer2016scalable,schafer2017multivariate} with strong ability of frequent extraction can significantly improve the performance of the TSC models. Thus, utilizing frequency analysis as an auxiliary tool to explore the learning behavior of 1D-CNNs is helpful to understand their strengths and limitations.

\begin{figure*}[ht]
    \centering
    \includegraphics[width=0.93\textwidth]{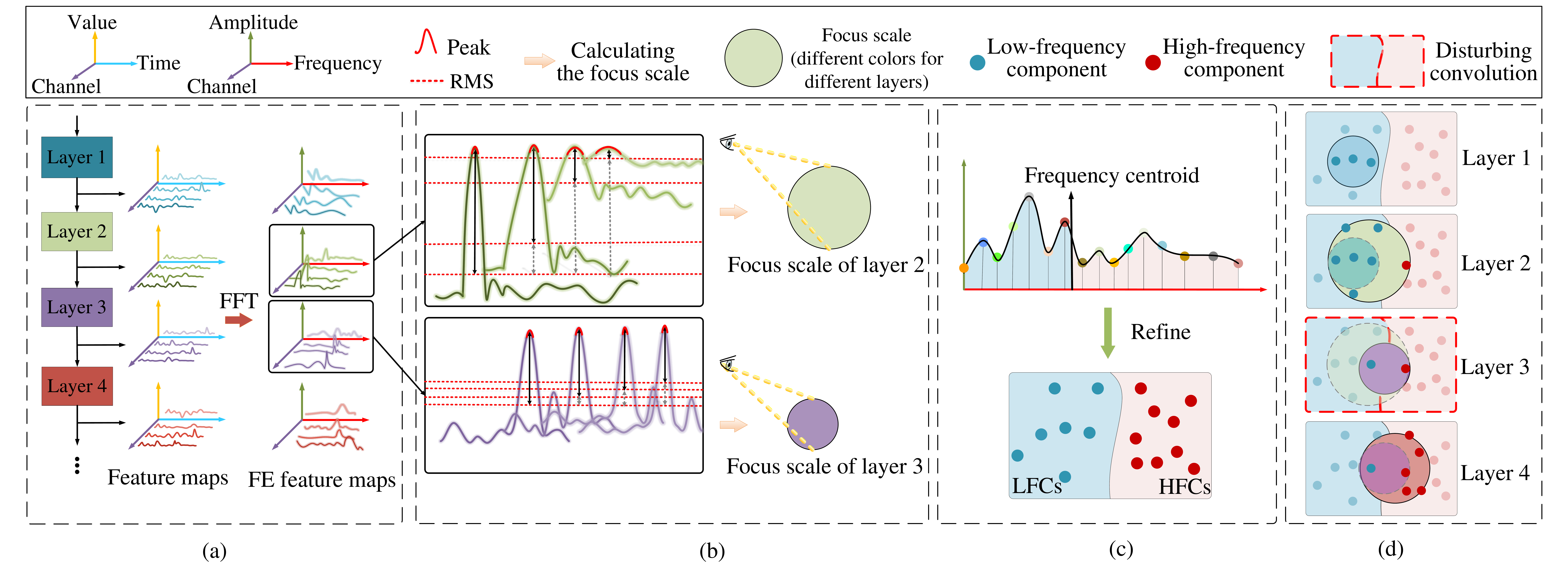} 
    \caption{Illustration of TCE in four convolutional layers. (a) Transform feature maps from the time domain to the frequency domain to obtain FE feature maps. (b) Get the focus scales from FE feature maps with layer 2 and layer 3 as examples. (c) Refine frequency components to HFCs and LFCs by the frequency centroid, demonstrated with an input instance variable as an example. Here, its frequency centroid being closer to 0 indicates that energy is concentrated at lower frequencies. (d) The learning behavior described in Remark \ref{re1}. Focus scales of layer 2 and layer 4 increase, while layer 3 (disturbing convolution) decreases. 
    With the shift of deeper focus to HFCs, layer 3 drives the network to distract from LFCs. }
    \label{fig_all}
\end{figure*}

More importantly, lots of works have proved that it is meaningful to explore the working mechanism of CNN capturing 2D images from the frequency perspective. \cite{xu2019training} introduced the frequency principle, elucidating CNN generalization, with subsequent in-depth theoretical proofs by \cite{xu2019frequency}. This principle clarifies that CNNs often fit target functions from low to high frequencies during the training process and high-frequency components are hard to encode. On this basis, CNNs were designed to fast learn a function with high frequencies to improve their performances \cite{zhang2020type,liu2020multi,jagtap2020adaptive}. In line with this, \cite{rahaman2019spectral} pointed out that the CNNs exhibit a bias towards low frequencies, and \cite{xu2021deep} explored the bias of deeper hidden layers towards lower frequencies during the initial stages of training. 
However, previous efforts primarily aimed at understanding CNNs in image data processing, which differs significantly from time series analysis. In contrast, we focus on addressing the complex and rarely explored challenge of accuracy degradation in TSC. From this perspective, we investigate the underlying mechanism of 1D-CNNs to better understand their learning behavior and harness their potential to overcome limitations in TSC tasks.

\section{Preliminaries}
\paragraph{Problem formulation.} TSC tasks can be represented as a set of $N$ times instances tuple: $\{ ({\textbf{{U}}^1},{y^1}),...,({\textbf{{U}}^N},{y^N})\} $, where $\textbf{{U}}$ denotes individual instance and $y \in \{ 1,...,Y\}$ is the corresponding label in $Y$ classes. Each instance $\textbf{{U}} =\{\textbf{u}_1,...,\textbf{{u}}_D\}\in \mathbb{R}^{D\times T} $ consists of $D$ variables with $T$ length. The goal of TSC tasks is to train a classifier that can assign each instance $\textbf{{U}}$ to its corresponding label $y$.

\paragraph{1D convolution.} We model the 1D-CNNs to process the above TSC tasks. For $l$-th convolutional layer (or convolutional block with a group of convolutional layers) of a 1D-CNN with $L$ depth, the trainable 1D convolutional kernels can be represented as $\textbf{{W}}_l \in \mathbb{R}^{C_{l-1}\times C_{l}\times k}$ with $C_{l-1}$ as the input channel axis, $C_{l}$ as the output channel axis and $k$ as the length axis. Then, the input of $l$-th layer can be represented as $\textbf{X}_l \in \mathbb{R}^{C_{l-1}\times H_{l-1}}$ with $H_{l-1}$ as the length axis. For the first layer, the input $\textbf{X}_1$ is the individual instance $\textbf{{U}}$ of the TSC task. The output of $l$-th layer can be represented as $\textbf{O}_l \in \mathbb{R}^{C_{l}\times H_{l}}$, which is also called feature maps. More specifically, we let $\textbf{x}_l^{c_{l-1}}$ and $\textbf{o}_l^{c_{l}}$ denote $c_{l-1}$-th channel input of $\textbf{X}_l$ and $c_{l}$-th channel feature map of $\textbf{O}_l$ respectively. $\textbf{w}_l^{c_{l-1},c_{l}}$ denotes the $c_{l-1}$-th input channel and $c_{l}$-th output channel convolutional kernel of $\textbf{W}_l$. Therefore, the 1D convolution can be defined as
\begin{equation}
    \label{conv_eq}
    \textbf{{o}}_l^{c_l} = \sum\limits_{c_{l-1} = 1}^{C_{l-1}} {\textbf{w}_l^{c_{l-1},c_{l}}\circledast \textbf{x}_l^{c_{l-1}}}+b_l^{c_l},
\end{equation}
where $\circledast$ denotes convolution operation, $b_l^{c_l}$ denotes the corresponding bias term.  The above 1D convolution operation is defined as $\mathcal{F}^{conv}(\cdot )$.

\section{Temporal Convolutional Explorer}
In this section, we introduce the \emph{Temporal Convolutional Explorer (TCE)} mechanism to uncover 1D-CNNs' learning behavior. We present Frequency-Extracted (FE) feature maps, offering a distinct representation  of frequency features extracted by convolutional kernels. Next, we propose the focus scale and frequency centroid concepts. Integrating these aspects, TCE identifies emphasized/overlooked frequency components in deeper layers, shedding light on the network's learning behavior. We reveal that deeper 1D-CNNs may contain disturbing convolutions that cause accuracy degradation.

\subsection{FE Feature Maps}
Feature maps are the representation of the learning features of convolution. To obtain its frequency features, TCE  transforms the feature maps by FFT and yields FE feature maps, which are employed as its basic feature maps, through the following definition.
\begin{definition}
    \label{def1}
    For any time sequence $\textbf{s}$ with $T$ time length ($\textbf{s}$ can be $\textbf{o}_l^{c_l}$, $\textbf{x}_l^{c_l}$ or $\textbf{u}_d$ in this paper), the transformation of $\textbf{s}$ from time domain to frequency domain can be described as
    \begin{displaymath}
        \textbf{f}^{\omega} = \frac{1}{T}\sum\nolimits_{t = 0}^{T-1} {\textbf{s}^{t}} {e^{ - j \omega t/T}}, \nonumber
    \end{displaymath}
    where  $\omega \in \{{ 0,...,\lfloor \frac{T-1}{2} \rfloor}\}\times  \frac{\omega_s}{T}$ denotes the frequency unit. $\omega_s$ denotes the sampling frequency. $\lfloor \cdot \rfloor$ is the floor bracket, which rounds the number to a lower integer. $\textbf{f}^{\omega}$ is a complex number. Let $\textbf{r}^\omega$ and $\textbf{i}^\omega$ represent the real part and imaginary part of $\textbf{f}^{\omega}$ respectively.
    We can get the amplitude at $\omega$:
    \begin{equation}
        \textbf{z}^\omega = \sqrt {(\textbf{r}^\omega)^2 + (\textbf{i}^\omega)^2}. \nonumber
    \end{equation}
    Therefore, we define the following transformation mapping:
    \begin{equation}
        \mathcal{A}: \textbf{s}\rightarrow \textbf{z}.\nonumber
    \end{equation}
\end{definition}
For the $c_l$-th feature map $\textbf{o}_l^{c_l}$ of $l$-th convolutional layer, we define $\mathcal{A}(\textbf{o}_l^{c_l})$ as its corresponding FE feature map. The above process is shown in Fig. \ref{fig_all} (a).

\subsection{Learning Behavior \& Disturbing Convolution}
With FE feature maps as the spectrum of the output feature maps, we further explore the learning behavior of each convolutional layer. We first use the ratio relationship between the peak and root-mean-square (RMS) of amplitude in the FE feature map as the representative indicator of the feature response. For $c_l$-th FE feature map $\mathcal{A}(\textbf{o}_l^{c_l})$ in $l$-th convolutional layer, it can be calculated by
\begin{equation}
    \label{peak}
    p_l^{c_l} = \frac{\max\limits_{\omega}\{\mathcal{A}^{\omega}(\textbf{o}_l^{c_l})\}}{{\sqrt {\sum\nolimits_{\omega\in \textbf{I}} {(\mathcal{A}^{\omega}(\textbf{o}_l^{c_l}))^2}}}}\times \sqrt{{\left\lfloor \frac{H{}_l-1}{2} \right\rfloor}+1},
\end{equation}
where $\textbf{I}=\{{ 0,...,\lfloor \frac{H_l-1}{2} \rfloor}\}\times  \frac{\omega_s}{H_l}$, $\mathcal{A}^{\omega}(\textbf{o}_l^{c_l})$ denotes the amplitude value of FE feature map $\mathcal{A}(\textbf{o}_l^{c_l})$ at the frequency $\omega$. 
The ratio calculated using Eq. \ref{peak} characterizes the concentration of the frequency response, which has been shown to accurately classify various signals \cite{paulter2004ieee,singh2017application}. 
By calculating this ratio for each FE feature map in a convolutional layer, we can effectively distinguish the frequency response characteristics within each channel. 

To assess the richness of frequency responses across feature maps within a convolutional layer, we introduce the concept of \emph{focus scale}. By calculating the variance of these patterns across all FE feature maps in the $l$-th convolutional layer, the focus scale $v{}_l$ of the $l$-th convolutional layer can be defined as 
\begin{equation}
    \label{var}
    {v{}_l} = \frac{{\sum\nolimits_{c_l = 1}^{C{}_l} {(p_l^{c_l} - \sum\nolimits_{c_l^{\prime } = 1}^{C{}_l} {p_l^{c_l^{\prime}}/C{}_l{)^2}} } }}{{C{}_l}}.
\end{equation}
We use the focus scale to evaluate the diversity of frequency patterns captured by each feature map within the $l$-th convolutional layer. 
A larger focus scale implies that the layer responds to a broader range of frequency features.
Thus, the focus scale ${v{}_l}$ provides an intuitive understanding of the frequency \emph{range} captured by the $l$-th convolutional layer. The process of computing the focus scale is displayed in Fig. \ref{fig_all} (b). 

To understand the frequency composition within the focus scale, we utilize the \emph{frequency centroid} of a signal to distinguish its high-frequency components (HFCs) and low-frequency components (LFCs), as depicted in Fig. \ref{fig_all}(c). The frequency centroid, denoted as $\mathcal{F}^{fc}(\textbf{s})$, represents the distribution centroid of the frequency components within a sequence $\textbf{s}$. It is formulated as
\begin{align}
    \label{fc}
    \mathcal{F}^{fc}(\textbf{s}) & = \frac{{\int_0^{\lfloor{(T-1)}/2\rfloor} {\omega \mathcal{A}^{\omega}(\textbf{s})\textbf{d}{\omega}} }}{{\int_0^{\lfloor{(T-1)}/2\rfloor} {\mathcal{A}^{\omega}(\textbf{s})\textbf{d}{\omega}} }},
\end{align}
where $\mathcal{A}(\cdot)$ is the transformation mapping in Definition \ref{def1}. 
Within the frequency spectrum of a signal, the frequency centroid acts as a measure of central tendency for the signal's frequency components. It facilitates the effective identification and analysis of spectral properties in diverse time series signals. Specifically, a frequency centroid located at the frequency center indicates a uniform frequency distribution, with no preference for lower or higher frequencies. What's more, a lower frequency centroid signifies a greater concentration of energy in the low-frequency range of the signal, highlighting significant information within that specific domain. Conversely, a higher frequency centroid emphasizes the presence of high-frequency information of the signal, associated with rapid changes or noise disturbances.

By considering the information regarding the frequency content and energy distribution, we can distinguish between LFCs and HFCs based on the relative positions of frequencies in relation to the centroid. This distinction, as illustrated in Fig. \ref{fig_all}(c), establishes a natural reference point for categorizing the frequency components into two distinct groups: LFCs located below the centroid and HFCs located above it. Through this reference, we can filter specific frequency components using the inverse FFT. This enables us to observe the learning behavior of the 1D-CNNs with respect to the targeted frequency components.

Moreover, monitoring the evolution of frequency centroids within the feature maps of each convolutional layer aids in comprehending the learning bias of individual layers concerning frequency components. When a deeper convolutional layer focuses its attention on modeling HFCs within the signal, as visually depicted in Fig. \ref{fig_all} (d), it triggers a concentrated activation of high-frequency energy, resulting in an elevated frequency centroid. Hence, the frequency centroid helps to understand the learning \emph{direction} of deeper 1D-CNNs in terms of frequency.

With the focus scale and the focus centroid, we can analyze the changes in focus with respect to both \emph{range} and \emph{direction}. It reflects the frequency components \emph{emphasized/overlooked} in the time series by deeper convolutional layers, as presented in Fig. \ref{fig_all} (d). The following remark clarifies the learning behavior of 1D-CNNs.

\begin{remark}
    1D-CNNs with the increase of depth tend to distract the focus from the LFCs. TCE can utilize the change in focus scales, which is defined by
    \begin{equation}
        {M}_l = { v}_l - { v}_{l-1}, \nonumber
    \end{equation}
    to describe the internal factor of this tendency: $l$-th convolutional layer with negative ${M}_l$ causes the 1D-CNN to lose focus on certain frequency components, mainly the LFCs.
    \label{re1}
\end{remark}
Remark \ref{re1} elucidates the inherent learning behavior of deeper 1D-CNNs, indicating the challenges in maintaining accuracy due to their distraction from low frequencies. Specifically, with the frequency centroid of TCE, we observe that LFCs significantly contribute to the overall generalizability performance of 1D-CNNs in TSC tasks, and that deeper 1D-CNNs exhibit stronger learning ability for LFCs. However, the accuracy degradation issue indicates that deeper 1D-CNNs fail to capitalize on this superior ability, as they distract from the focus on low frequencies that are crucial to their performance.
The observations form the basis of Remark \ref{re1}. For example, layer 3 in Fig. \ref{fig_all} (d) displays a narrowing range of captured frequency features, which prevents them from fully exploiting the generalizable low-frequency features, ultimately resulting in accuracy degradation. We refer to the convolutional layer with negative ${M_l}$ as the \emph{disturbing convolution}. The aforementioned conclusions will be further substantiated in Sec. \ref{sec_exp} through empirical studies conducted using real-world data.
\section{Regulatory Framework} 
\begin{figure}[t]
  \centering
  \includegraphics[width=0.88\columnwidth]{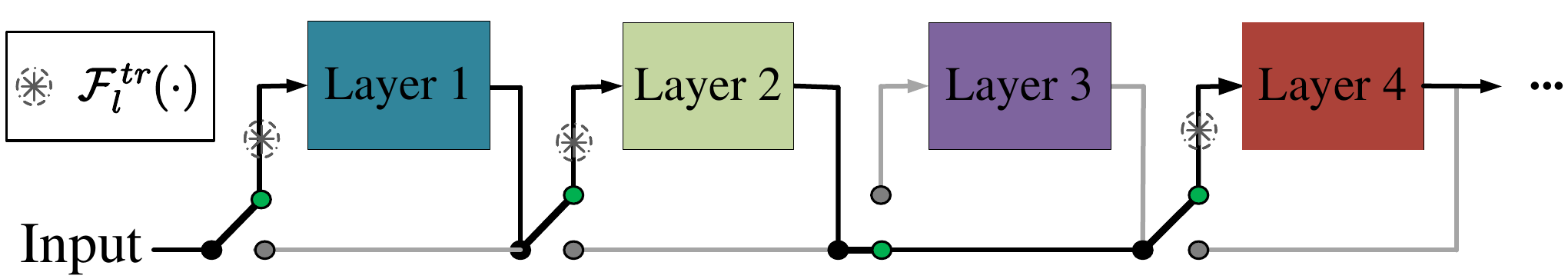} 
  \caption{Illustration of gating mechanism. Layer 3 is skipped over. $\mathcal{F}_l^{tr}(\cdot)$ can ensure the consistent channel dimension.
  }
  \label{1_2}
\end{figure}

To leverage our findings of 1D-CNNs' learning behavior to a practical application, we propose a regulatory framework to rectify the suboptimal learning behavior and alleviate the accuracy degradation by enabling the network to selectively bypass the specified disturbing convolutions.
\paragraph{Gating mechanism.}
We first introduce our gating mechanism to skip over any specified convolutional layers. As shown in Fig. \ref{1_2}, the gating mechanism adds a switch structure to the original network making each convolutional layer have two different states: skipped or preserved. In the gating mechanism, the input of $(l+1)$-th convolutional layer is redefined as
\begin{equation}
    \label{conv'}
    {\textbf{X}_{l+1}^\prime} = g_l\times{\rm ReLU}(\mathcal{F}^{conv}_l (\mathcal{F}^{tr}_l(\textbf{X}_l^\prime))) + (1 - g_l)\times \textbf{X}_l^\prime,
\end{equation}
where ${\rm ReLU}(\cdot)$ is the activation function, $g_l \in \{0,1\}$ is the gating factor. $g_l=0$ means that $l$-th convolutional layer is skipped, $g_l=1$ means that $l$-th convolutional layer is preserved.
However, the channel dimension of ${\textbf{X}_{l}^\prime}$ may not match the input channel dimension of $l$-th convolution in the gating mechanism. 
Therefore, $\mathcal{F}^{tr}_l(\cdot)$ is used to enable $\textbf{X}_l^\prime$ applied in $\mathcal{F}^{conv}_l(\cdot)$ with consistent dimension. If the channel dimensions are matched, $\mathcal{F}^{tr}_l(\textbf{X}_l^\prime)=\textbf{X}_l^\prime$, otherwise, $\mathcal{F}^{tr}_l(\textbf{X}_l^\prime)={\rm ReLU}(\textbf{W}^{\prime}_l \circledast \textbf{X}_l^\prime)$, where $\textbf{W}^{\prime}_l$ is the weight of an $l$-length convolution to realize the channel transformation.
\paragraph{The overall framework.}
Remark \ref {re1} naturally motivates us to propose a gating solution to bypass disturbing convolutions. We utilize $\textbf{M}=\{M_1,...,M_L\}$ in TCE to determine the gating factor $\textbf{g}=\{g_1,...,g_L\}$ with the following steps. 1) Identify all disturbing convolutions with the negative ${M}_l$. 2) Sort these disturbing convolutions according to ${M}_l$ in ascending order. 3) Select the first $\mathcal{P}$ disturbing convolutions to form the set $\textbf{M}^{\ast }$ as the object set that will be skipped over. 4) Assign $g_l = 0$ for elements in $\textbf{M}^{\ast}$ and $g_l = 1$ for elements in $\textbf{M} - \textbf{M}^{\ast}$. By incorporating TCE and the gating mechanism, we propose a plug-and-play regulatory framework that enables the network to selectively bypass the specified disturbing convolutions with the gating factor $\textbf{g}$.

After implementing our regulatory framework during the $\alpha$-th epoch of the training process, the refined regulated network is then applied in the remaining training process and the testing process. Specifically, the regulatory framework divides the entire training process of an original network into two stages. During the first $\alpha$ epochs, we train the original network as usual, and run once above steps on the last batch of training samples at the end of $\alpha$-th epoch to calculate the gating factor $\textbf{g}$. During the remaining epochs, we train the regulated network after skipping disturbing convolutions, loading the corresponding parameters of the original network at the beginning of $(\alpha + 1)$-th epoch. These two stages will not change the number of original training epochs, and the regulated network that has fewer parameters and requires less computation can also benefit from spending less training time. We apply the regulated network to the testing process, which can effectively alleviate the negative effects of disturbing convolutions as well as can reduce the consumption of memory and computational overhead. 
Additionally, in Sec. \ref{sen}, we offer observations and insights regarding the selection of two hyperparameters within the regulatory framework, i.e.,  $\alpha$ and $\mathcal{P}$.

\section{Experiment} \label{sec_exp}
\subsection{Experimental Details}
\subsubsection{Datasets.} 
To verify the generality of our findings, we conduct experiments on a wide range of datasets \footnote{\begin{minipage}{1\linewidth}Datasets and the descriptions are at \url{http://timeseriesclassification.com} and \url{http://archive.ics.uci.edu/ml/datasets/Human+Activity+Recognition+Using+Smartphones}.\end{minipage}}. The details of each kind of dataset are as follows.
\begin{itemize}
  \item \textbf{128 UTSC Datasets}. A collection of 128 UTSC datasets is sourced from the UCR archive \cite{dau2019ucr}. These datasets encompass various domains (e.g., health monitoring,  remote sensing and speech recognition) and exhibit distinct characteristics in terms of size and length.
  \item \textbf{31 MTSC Datasets}. 1) 30 MTSC datasets from the UEA archive  \cite{bagnall2018uea} encompass diverse application domains (e.g., ECG and motion classification) and differ in terms of the number of channels, size, and length. 2) One large-scale MTSC dataset from the UCI archive \cite{asuncion2007uci}, namely the Human Activity Recognition (HAR) dataset, consists of 10,299 instances with 6 categories and 9 channel variates. The processing of this HAR dataset follows the same approach as described in \cite{karim2019multivariate}.

\end{itemize}
To verify the learning behavior of 1D-CNNs proposed by TCE, we focus on two UTSC tasks (CricketX and FaceAll) from the UCR archive, as well as two MTSC tasks (SelfRegulationSCP1 and HAR) from the UEA/UCI archives. These datasets represent various types of data (e.g., motion, image, and health), and range in size from 268 to 10,299 instances. Importantly, these four datasets share a common spectral property inherent in time series signals, where low frequencies dominate the spectrum. 
To be specific, we have employed Eq. \ref{fc} to calculate the instance-wise and variable-wise frequency centroids for all datasets within the aforementioned three archives, encompassing 128 UTSC datasets and 31 MTSC datasets. The ratio of the calculated frequency centroids in relation to their respective maximum frequency exhibits an average value of 0.26, with a corresponding variance of 0.01. The proximity of this average ratio to zero suggests that lower frequencies carry a substantial amount of energy and important information in the majority of time series signals. Thus, we select the four datasets whose ratio values closely align with the aforementioned average value. This selection indicates that these four datasets effectively represent typical frequency patterns encountered in various types of time series data. By focusing on these datasets, we can intuitively explore the learning behavior of 1D-CNNs on LFCs and significantly enhance the generalizability of our experimental findings. Moreover, all datasets of the three archives (including 128 UTSC datasets and 31 MTSC datasets) will be applied to evaluate the performance of our regulatory framework. 

\subsubsection{Experiment setup.}
To verify the effectiveness of TCE as a general exploration method for 1D-CNNs, we conduct experiments using three competitive 1D-CNN backbones, following the recent empirical surveys on TSC methods \cite{wang2017time,ruiz2021great}. Specifically,
FCN \cite{wang2017time}, ResNet \cite{wang2017time} and InceptionTime \cite{ismail2020inceptiontime} are applied to UTSC tasks, and ResNet \cite{ruiz2021great} and InceptionTime \cite{ruiz2021great} are applied to MTSC tasks. For FCN's adaptation to MTSC tasks, we adjust its input channel count to align with the number of input sample channels. Residual/Convolutional blocks are sequenced in ResNet and InceptionTime, while FCN comprises convolutional layers. Network depth is indicated by the number of blocks or layers.
Models are trained by the Adam optimizer \cite{kingma2014adam} with learning rate set to be from $10^{-6}$ to $10^{-3}$. Consistent with the original papers, we adopt cross-entropy loss as the loss function, employ a batch size of 16, conduct training for 1500 epochs, and initialize weights through Xavier initialization \cite{glorot2010understanding}.  Our code can be found at this link\footnote{\url{https://github.com/jrzhang33/TCE}\label{footlabel}}.

To identify networks that suffer from significant accuracy degradation issues, we conduct a comparative analysis between deeper 1D-CNNs and shallower 1D-CNNs. If the accuracy of a deeper network is found to be at least 5\% lower than that of a shallower network, we classify the deeper network as experiencing accuracy degradation. To corroborate the insights provided by TCE regarding accuracy degradation in deeper 1D-CNNs, we focus our in-depth investigation on ResNet with a depth of 5, referred to as \emph{deeper ResNet}. These networks consistently exhibit accuracy degradation issues across four representative datasets, as depicted in Fig. \ref{basic_fig}. 
In our regulatory framework, we set $\alpha = 100$ and $\mathcal{P} = 2$. The regulator is then applied to all three backbone networks, aiming to assess the generality and effectiveness of the regulatory framework in addressing the accuracy degradation observed in deeper 1D-CNNs. Finally, we conduct a sensitivity analysis to explore the impact of different settings for the two hyperparameters, $\alpha$ and $\mathcal{P}$, which are unique to the regulatory framework, on the classification accuracy. 

\subsubsection{Grad-CAM criteria.}
In the next experimental analysis, with Gradient-weighted Class Activation Mapping (Grad-CAM) \cite{2020Grad}, we identify the contribution of each temporal region to the network's output class $y$ for a given time series instance $\textbf{U}$.
We first compute the gradient of the score $\lambda^y$ (before the softmax) for the class $y$, concerning $c_L$-th channel output ${\rm ReLU}(\textbf{{o}}_L^{c_L})$ of the last convolutional layer: ${\partial{\lambda ^y}}/{{\rm ReLU}(\textbf{{o}}_L^{c_L})}$. Then, calculate the importance weights of $c_L$-th channel output in the last convolutional layer by
\begin{equation}
    \alpha_{c_L}^{\lambda^y} = \frac{1}{H_L}\sum\nolimits_{t=1}^{H_L}{\frac{\partial \lambda^y}{\partial {\rm ReLU}(\textbf{o}_L^{c_L,t})}}.
\end{equation}
Finally, a weighted combination is performed followed by a ReLU activation:
\begin{equation}
    \textbf{E}^{\lambda^y} = {\rm ReLU}\left(\sum\nolimits_{c_L=1}^{C_L}{\alpha_{c_L}^{\lambda^y}{\rm ReLU}(\textbf{o}_L^{c_L})}\right),
\end{equation}
where ${\textbf{E}^{\lambda ^y} \in \mathbb{R} ^T}$ is the Grad-CAM contribution and $E_t^{\lambda ^y}$ is the activation value for class $y$ at time $t$, which expresses the contribution of input at time $t$ for the output class in instance $\textbf{U}$. Therefore, with Grad-CAM, we can identify the contribution of each temporal region to the output classification.

\subsection{Absence of Overfitting.} \label{secoverfitting}
\begin{figure}[tb]
  \centering
  \includegraphics[width=0.9\columnwidth]{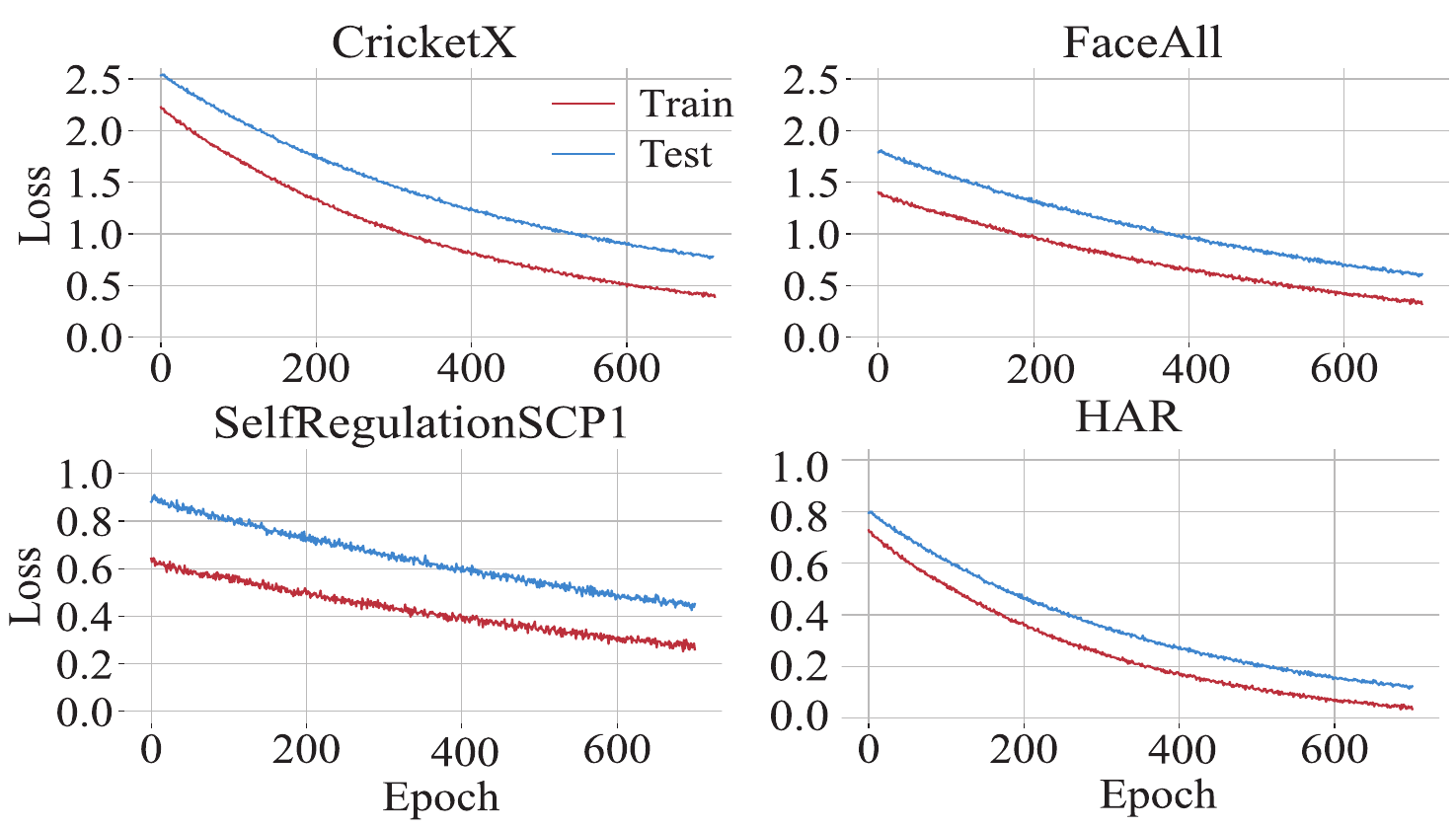} 
  \caption{Traning and test loss curves of deeper ResNet on four TSC datasets.}
  \label{overfitting}
\end{figure} 
\begin{figure}[tb]
  \centering
  \includegraphics[width=0.8\columnwidth]{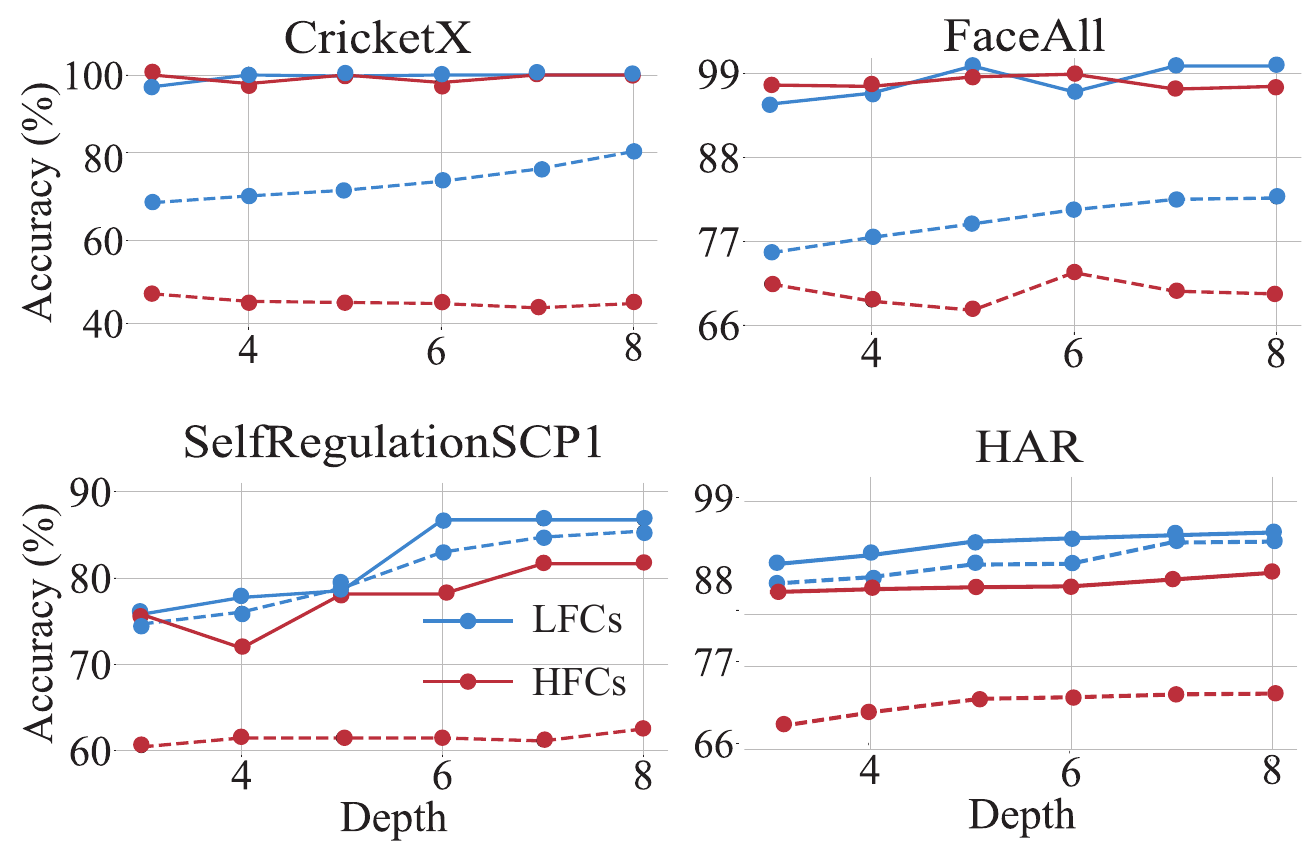} 
  \caption{The training and test accuracies of ResNet at different depths on LFCs and HFCs. The accuracy curves of LFCs and HFCs are displayed in blue and red respectively. Solid lines and dotted lines denote training and test accuracy curves respectively.}
  \label{lowhigh}
\end{figure} 

To investigate the accuracy degradation observed in the deeper ResNet, we first monitor the loss curve of this network to rule out the possibility of overfitting as the cause. Fig. \ref{overfitting} depicts the training and test loss curves for the deeper ResNet on the four datasets, consistently demonstrating a downward trend. This consistent decline in the curves suggests that deeper ResNet does not overfit to high-frequency noise, and as such, overfitting is not the fundamental factor contributing to the accuracy degradation.

\subsection{Learning Ability for LFCs.}\label{sec.}
In order to explore the generalization performance of the networks on different frequency components, we conduct a quantitative analysis using the frequency centroid of TCE. This involves filtering out specific frequency components in four datasets individually \footnote{We utilize FFT to transform the input time series into the frequency domain, setting the amplitudes of LFCs/HFCs to zero. Subsequently, we apply the inverse FFT to obtain time-domain signals containing only HFCs/LFCs.}. 
In Fig. \ref{lowhigh}, a clear result is observed that as the network depth increases, the difference between the training and test accuracy is consistently noticeable when learning HFCs. However, in the case of LFCs, the difference is less prominent and tends to decrease even further.
This observation suggests that 1D-CNNs struggle to generalize well on HFCs, even with increased depth. 
Thus, the learning behavior on HFCs has a limited impact on the overall performance of deeper networks.
Conversely, \emph{LFCs significantly contribute to the generalization performance of 1D-CNNs in TSC tasks}, and deeper networks are better equipped to leverage and generalize LFCs. This underscores that the deeper networks' learning on LFCs plays a crucial role in determining their overall performance on TSC tasks.

\begin{figure}[tb]
  \centering
  \includegraphics[width=0.87\columnwidth]{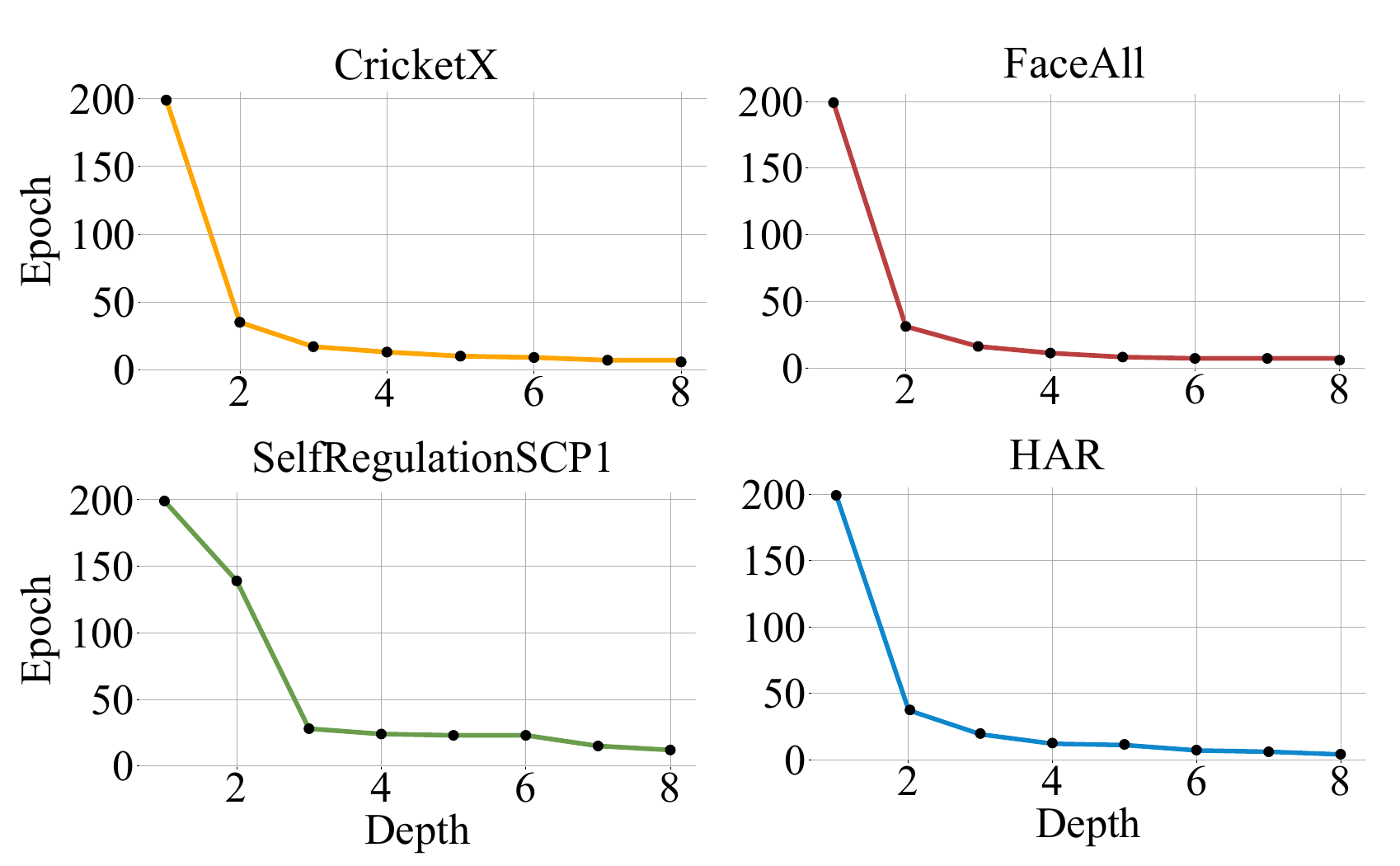} 
  \caption{The required training epochs for ResNets with different depths to achieve the target loss on LFCs.}
  \label{epoch}
\end{figure} 
\begin{figure}[tb]
  \centering
  \includegraphics[width=0.8\columnwidth,height=0.4\linewidth]{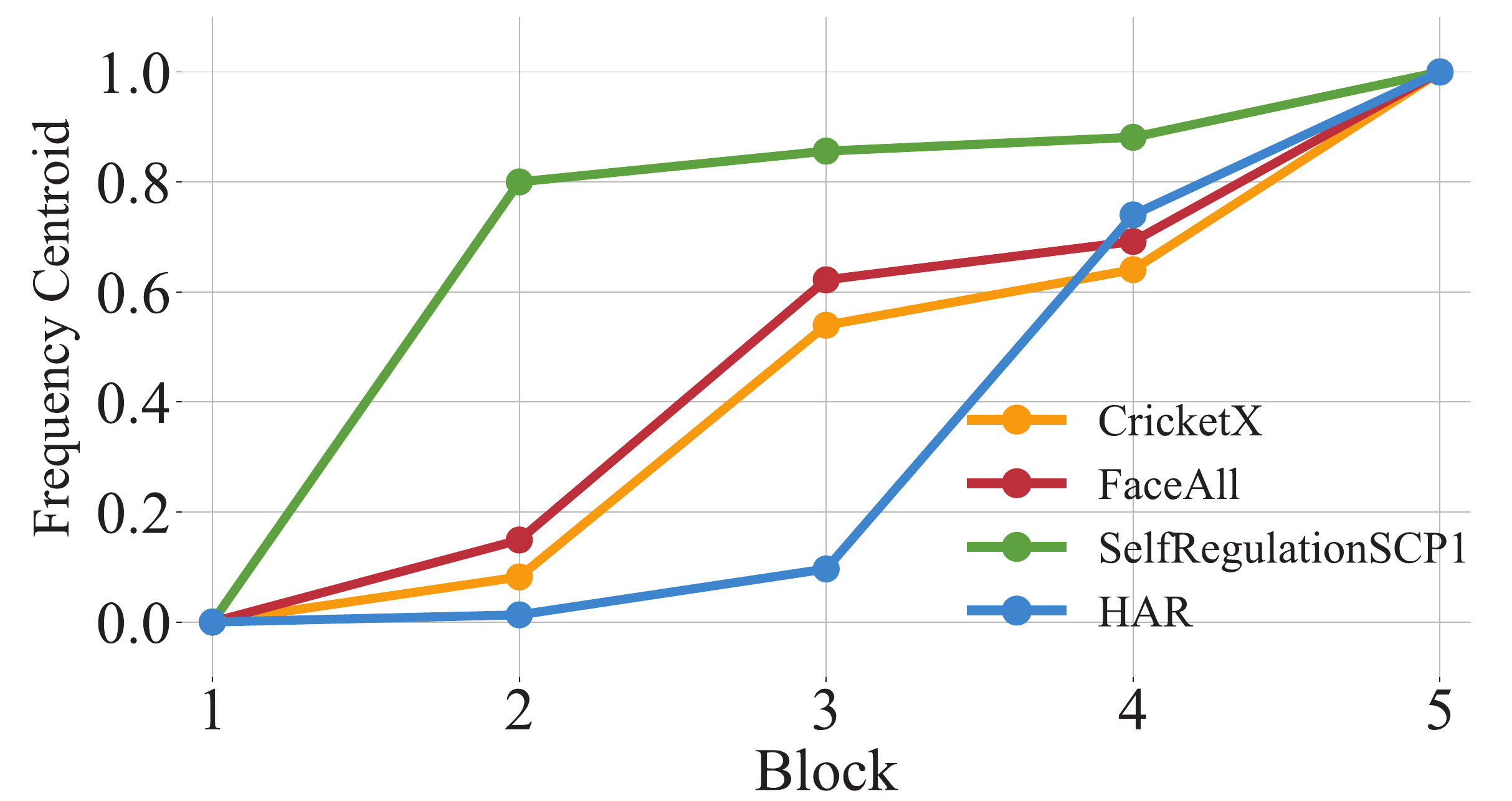} 
  \caption{The average frequency centroid of feature maps for each convolutional block in deeper ResNet. The frequency centroids within each dataset are scaled to the range [0, 1].}
  \label{fc_ver}
\end{figure} 

To further assess the learning ability of deeper networks on LFCs, we define that 1) the learning speed of CNN is faster if the loss of the network decreases to a target loss with fewer training epochs, and 2) the learning ability of CNN is stronger if the network generalizes well and learns faster. Fig. \ref{epoch} shows the learning speed of ResNet at different depths on LFCs. The results show that ResNet with more blocks achieves the target training loss with fewer epochs, which demonstrates that deeper networks have faster learning speeds for LFCs. Combined with the good generalization on LFCs as shown in Fig. \ref{lowhigh}, we can conclude that \emph{deeper 1D-CNNs exhibit stronger learning ability for LFCs in time series data}.

\subsection{Analysis on Learning Behavior.}
In this subsection, we will find out why such stronger learning ability for highly generalizable LFCs has not improved the performances of the deeper ResNet. First, we explore the learning direction of deeper ResNet in the frequency domain. Fig. \ref{fc_ver} showcases the frequency centroid of all feature maps in each convolutional block of the deeper ResNet. The results clearly demonstrate that the deeper layers exhibit significantly elevated frequency centroids, indicating a more pronounced activation of high-frequency energy. This analysis suggests that as the time series data progresses through successive layers, the network progressively focuses its attention on capturing higher frequency components. The visual representation of this evolving frequency centroid is presented in Fig. \ref{fig_all} (d), which distinctly highlights the shift in focus of deeper convolutional layers towards higher frequencies. Thus, there is a tendency for the deeper layers to overlook the presence of LFCs that are crucial to the network's performance in TSC tasks.
\begin{figure}[tb]
  \centering
  \includegraphics[width=0.85\columnwidth,height=0.55\linewidth]{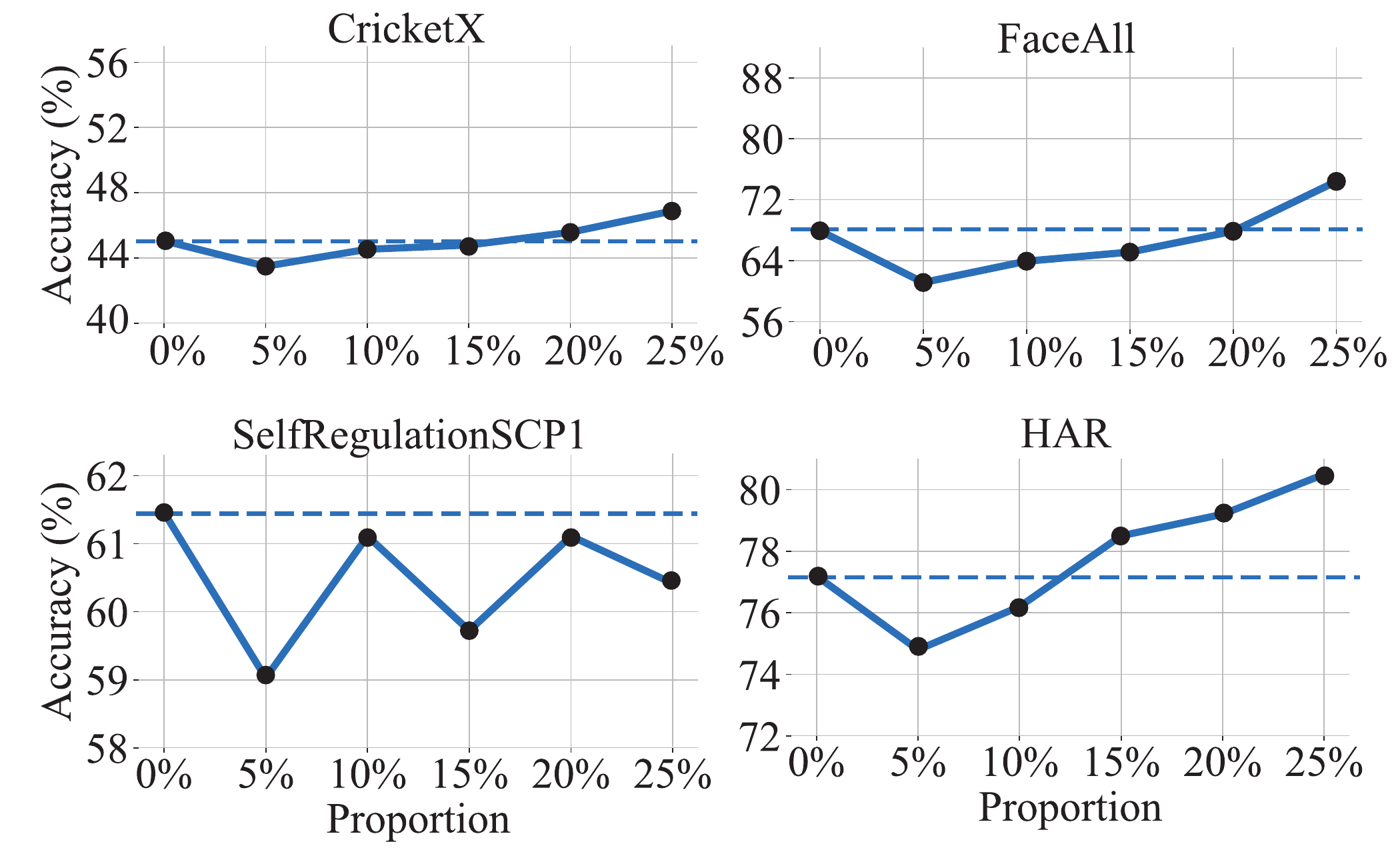} 
  \caption{Accuracy of deeper ResNet on four datasets with different proportions of LFCs added to HFCs.}
  \label{low}
\end{figure} 
\begin{figure}[tb]
\centering
\includegraphics[width=0.92\columnwidth,height=0.45\linewidth]{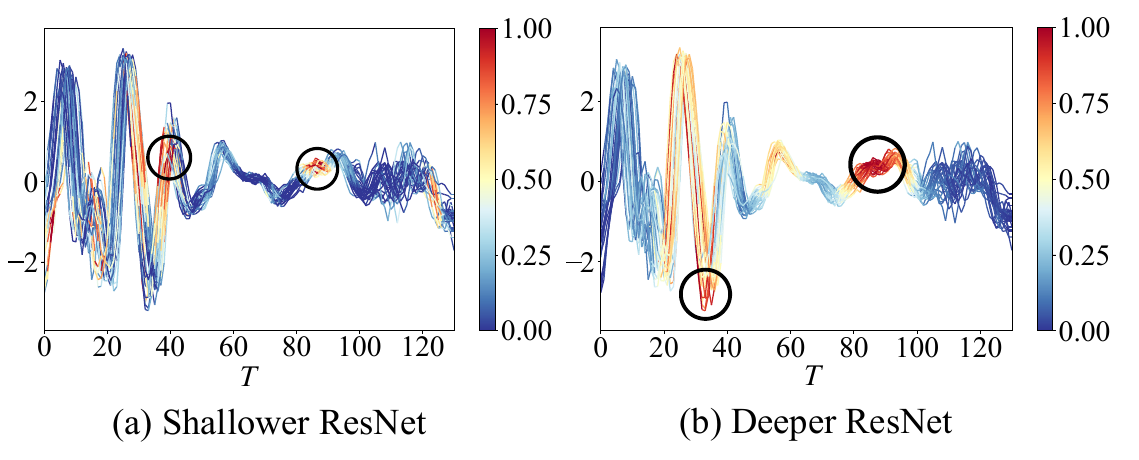} 
\caption{Grad-CAM of shallower and deeper ResNets on FaceAll instances in class 3.}
\label{cam}
\end{figure} 

  In light of this observation, we further concentrate on exploring the deeper ResNet's learning behavior on LFCs.
  To be specific, from TCE's frequency centroid towards the lower frequency, we gradually add (5\%, 10\%, 15\%, 20\%, 25\%) proportion of the LFCs to HFCs \footnote{We restore the amplitude of LFCs with various proportions from the frequency centroid of a frequency-domain signal that solely consists of HFCs. Subsequently, the time-domain signal with the desired proportions of LFCs can be obtained by applying the inverse FFT to that frequency-domain signal.}. As shown in Fig. \ref{low}, 
  when adding LFCs to HFCs of four datasets, the test accuracy does not gradually improve as expected, even lower than the initial accuracy on HFCs alone. In particular, in the SelfRegulationSCP1 dataset, the performance of deeper ResNet is continuously damaged by additional LFCs. 
  These results suggest that the deeper ResNet is unable to fully leverage its learning ability to generalize the low frequencies, which are restored from signals containing HFCs.
  Therefore, Fig. \ref{low} demonstrates that the presence of significant HFCs can impede the deeper ResNet from leveraging the generalization of the low-frequency information. In other words, the deeper ResNet's focus on capturing LFCs of original signals is disturbed by the existence of HFCs, preventing it from taking advantage of its learning ability for LFCs.

\begin{figure}[tb]
    \centering
    \includegraphics[width=1\columnwidth]{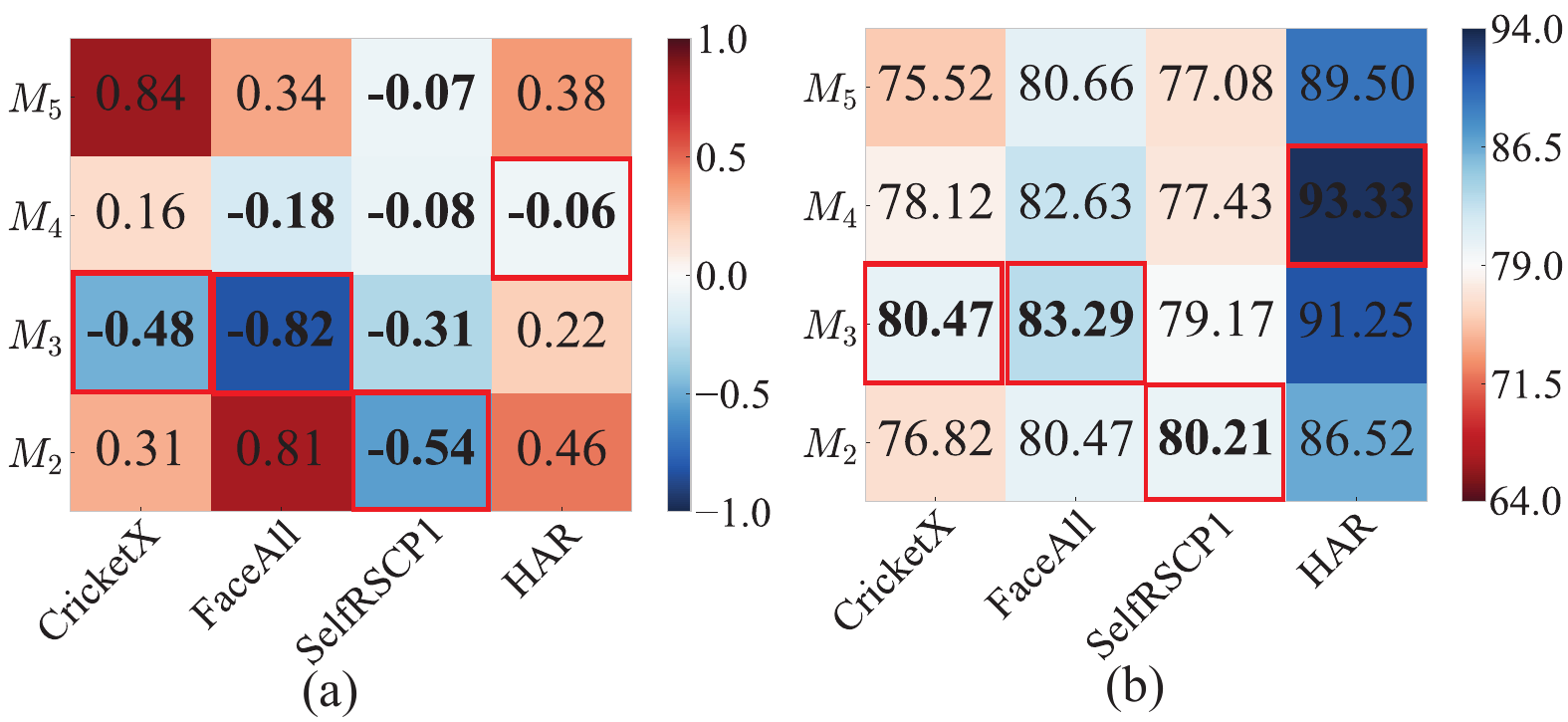} 
    \caption{Visualization and analysis on every block of deeper ResNet. (a) The changes in focus scales between blocks, normalized to the [0, 1] range. Negative changes are highlighted in bold, with the most negative change for each dataset framed in red. (b) Accuracy (\%) of networks after skipping each block, with the best results for each dataset highlighted in bold and framed in red.}
    \label{scale}
\end{figure} 

Finally, to furnish a more convincing substantiation of the distraction on capturing LFCs, we use Grad-CAM scores to compare the discriminative region difference between shallower ResNet with one convolutional block and deeper ResNet,  which shows worse performances on FaceAll instances in class 3. As indicated in Fig. \ref{cam}, shallower and deeper ResNets highlight different discriminative frequency features that contribute most to the predicting results. The shallower ResNet focuses on discriminative regions with slight fluctuations, while the deeper ResNet is limited to regions with shorter wavelengths or rapid oscillations. Therefore, the deeper network has shifted its focus towards fitting challenging HFCs, which confuses the extraction of generalizable low-frequency patterns, damaging the performance. By synthesizing these two analyses, we confirm the Remark \ref{re1}, that is, \emph{deeper 1D-CNNs distract the focus from low frequencies leading to the accuracy degradation issue}, even though they exhibit strong learning ability for LFCs.

\subsection{Analysis on Disturbing Convolution}
We visualize the changes in focus scales (i.e., $M_l$) and color the convolutional block with negative $M_l$ (i,e., disturbing convolution) \emph{blue} in deeper ResNets. From the results of each deeper ResNet for these four datasets in Fig. \ref{scale}(a), the presence of blue blocks is notably apparent, especially in the SelfRegulationSCP1 dataset. In accordance with the analysis of distraction on LFCs in Fig. \ref{low}, there are convolutional blocks with negative $M_l$ in deeper ResNet, which has been previously illustrated to suffer from accuracy degradation due to the loss of expression on LFCs.

Then, to confirm the relationship between disturbing convolution and the accuracy degradation, we skip each block in turn respectively by the gating mechanism and compare their accuracies in Fig. \ref{scale}(b). We find that the accuracy of every network after skipping the blue convolutional block with focus scale reduction consistently surpasses that of bypassing other blocks. This indicates that the convolutional block with a reduced focus scale has a more negative impact on accuracy, confirming that it possesses a narrower frequency response range.
By bypassing the disturbing convolutions with the most negative changes in focus scales, the accuracies of CricketX, FaceAll, SelfRegulationSCP1, and HAR exhibit the greatest improvements. The improvements are 8.86\%, 3.90\%, 6.60\%, and 4.97\% compared to Fig. \ref{basic_fig} respectively, which reveals that the disturbing convolution is the cause of accuracy degradation and skipping it can mitigate this issue.
\begin{figure}[tb]
    \centering
    \includegraphics[width=1\columnwidth,height=0.55\linewidth]{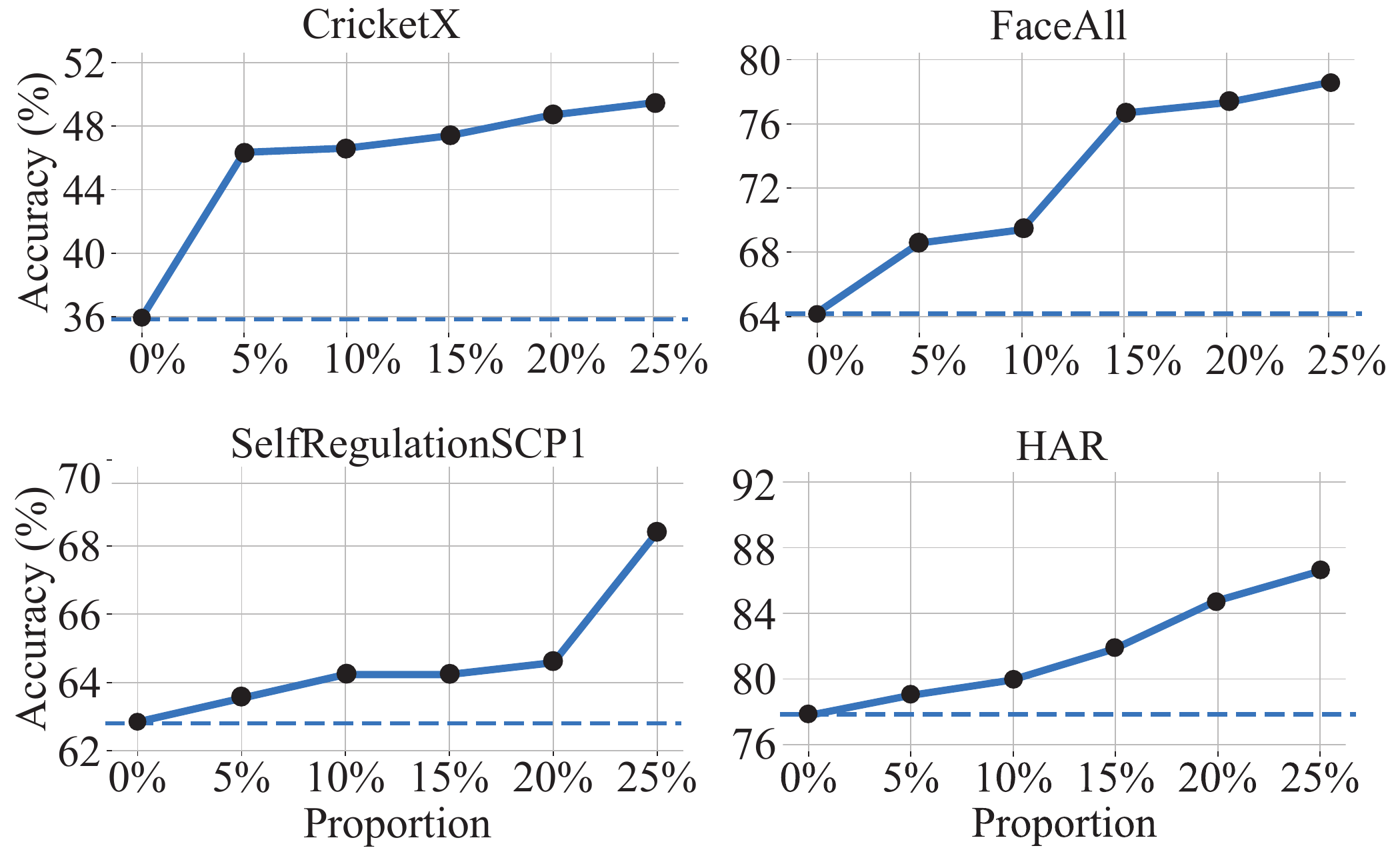} 
    \caption{Accuracy of deeper ResNet on four datasets with different proportions of LFCs added to HFCs, after skipping the specified disturbing convolution.}
    \label{low2}
\end{figure}

Finally, we bypass up to two disturbing convolutions and repeat the experiment where we add 5\% to 25\% proportion of the LFCs. As demonstrated in Fig. \ref{low2}, when adding LFCs to HFCs, the accuracies gradually improve. The result holds that deeper ResNet can refocus on the LFCs by skipping disturbing convolutions, and its learning ability for LFCs can be effectively leveraged for accuracy improvement. It also points out that the added LFCs contain valuable information for classification and skipping over disturbing convolutions can correct the suboptimal learning behavior.
Consequently, these results verify TCE's insight that \emph{the disturbing convolution is the potential driving factor for the accuracy degradation phenomenon caused by distraction on LFCs, and skipping over it can alleviate this issue.}

\subsection{Performances of Our Regulatory Framework}
Our regulatory framework is applied to FCN, ResNet, and InceptionTime (IT) to verify its performance. For quantitative evaluation, as done in \cite{2019Deep,dempster2021minirocket}, we conduct the pairwise posthoc analysis \cite{benavoli2016should} that statistically ranks different models according to their accuracies over all datasets. We add the analysis of parameters (Params) and floating point operations (FLOPs) as the measure of computational efficiency. We visualize the results by the critical difference (CD) diagram \cite{demvsar2006statistical} with Holm’s $\alpha$ (5\%) \cite{holm1979simple}. The diagram illustrates the average ranking of each classifier. The thick horizontal lines connect a group of networks that do not exhibit a significant difference in accuracy or computational efficiency. We repeatedly train each classifier with three different seeds and report the median results. 
\begin{figure}[tb]
  \centering
  \includegraphics[width=0.83\columnwidth]{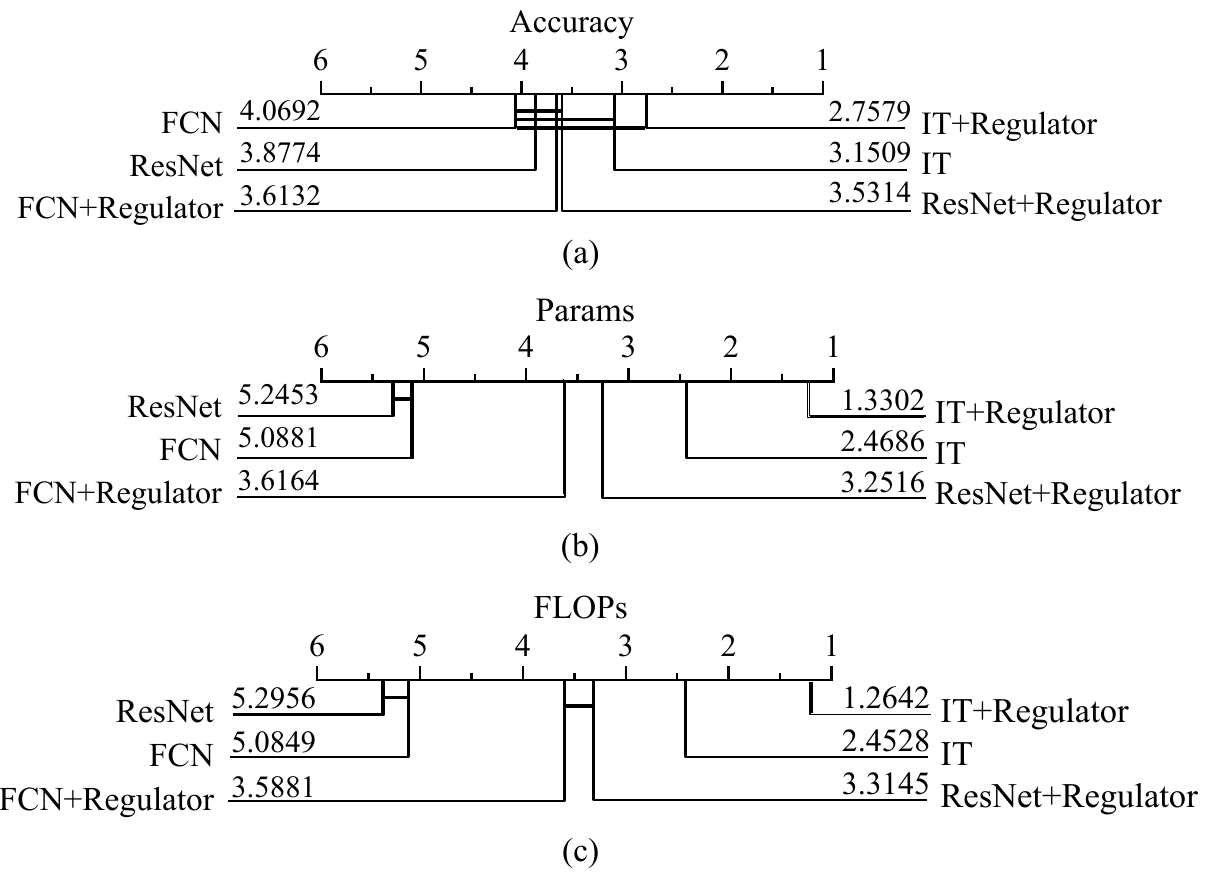} 
  \caption{CD diagrams comparing the performance of deeper 1D-CNNs with or without our regulatory framework (i.e., Regulator) in terms of (a) Accuracy, (b) Params and (c) FLOPs.}
  \label{result}
\end{figure} 

Fig. \ref{result} presents their average ranks over all datasets with the pairwise statistical differences. We observe that existing 1D-CNNs equipped with our regulatory framework exhibit better average ranks across all indicators (i.e. Accuracy, Params, and FLOPs) than their non-equipped counterparts, and the performance differences in terms of parameters and FLOPs are statistically significant. Therefore, our regulatory framework enables existing 1D-CNNs to significantly reduce the consumption of memory and computation resources while maintaining comparable or even higher accuracy. The successful application supports that TCE can provide hints for enhancing the performance of 1D-CNNs on TSC tasks. 

\subsection{Sensitivity Analysis of Our Regulatory Framework} \label{sen}
We perform a comparative analysis to evaluate the accuracy of the default configuration in relation to two different hyperparameters choices within our regulatory framework: $\alpha$ and $\mathcal{P}$. The hyperparameter $\alpha$ represents the epoch of network regulation, while $\mathcal{P}$ corresponds to the maximum number of disturbing convolutional layers that can be skipped.

\begin{figure}[t]
  \centering
  \includegraphics[width=0.93\columnwidth]{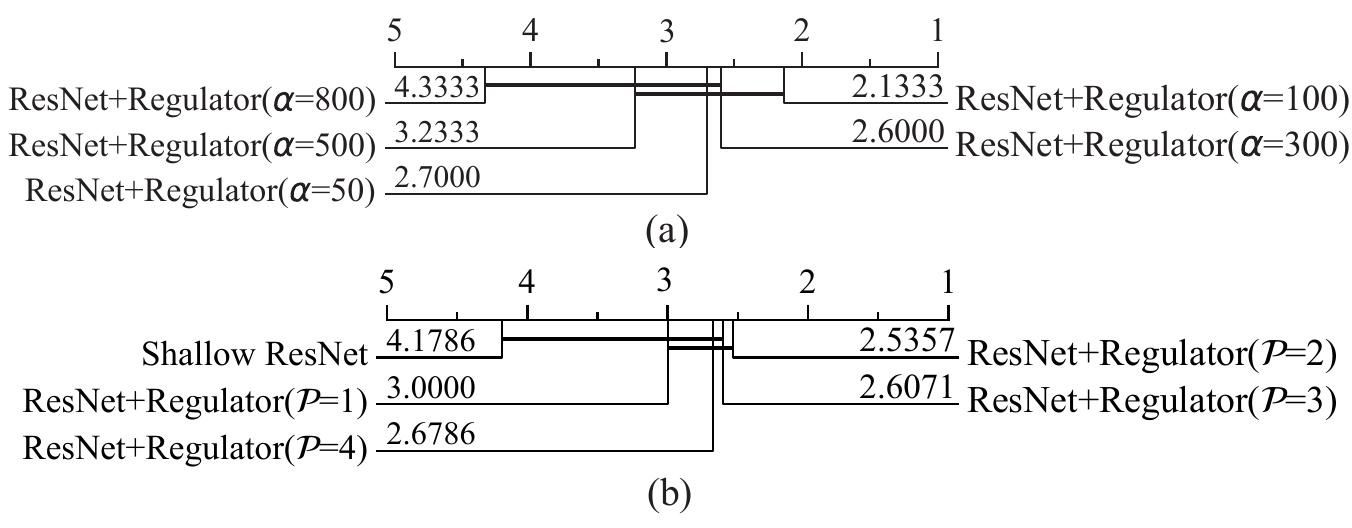} 
  \caption{CD diagrams comparing accuracy for different (a) $\alpha$ and (b) $\mathcal{P}$ values on deeper ResNet with our regulator.}
  \label{cd_sen}
\end{figure}

Fig. \ref{cd_sen} (a) shows the impact of the hyperparameter $\alpha$ on the accuracy of the regulator. The results indicate that for the majority of $\alpha$ values, the achieved accuracy remains relatively stable and does not deviate significantly from the best accuracy attained when $\alpha = 100$. A marked discrepancy in performance emerges only when $\alpha$ exceeds the threshold of 800. These findings suggest that the influence of the disturbing convolution stabilizes after approximately 100 epochs of network training, signifying a gradual loss of focus on LFCs by 1D-CNNs. By promptly applying our regulatory framework at this critical epoch, we effectively correct this suboptimal learning behavior, leading to enhanced network performance. When $\alpha \geq 800$, the performance improvement becomes impeded due to training saturation. This saturation effect indicates that the network has reached a plateau in its learning behavior, and further increases in the value of $\alpha$ do not yield substantial accuracy improvements. Thus, setting $\alpha$ to a value within a reasonable range, such as 100, proves sufficient to achieve performance gains while avoiding unnecessary training overhead.

Fig. \ref{cd_sen} (b) presents the effect of all the desirable values for another hyperparameter $\mathcal{P}$ (ranging from 1 to 4) on accuracy. In this comparative experiment, a shallow ResNet (depth = 2) without the regulatory framework is included for reference.
As observed in Fig. \ref{cd_sen} (b), there are no significant differences in performance when $\mathcal{P}$ takes values from 1 to 4. In contrast, the shallow ResNet without the regulator consistently exhibits the lowest performance in this case. These findings first indicate that the regulatory framework possesses robust adaptability to different settings of $\mathcal{P}$. Particularly, setting $\mathcal{P}$ to 2 yields optimal results as it strikes a fine balance between leveraging the learning abilities of deep networks and mitigating the adverse effects of convolutions.
Furthermore, the inferior performance of the shallow ResNet underscores the difficulty in adapting to diverse time-series data due to its limited learning ability and inherent learning behavior. This underscores the importance of integrating TCE's insights into the regulation of 1D-CNNs, which not only preserves the enhanced ability of deep 1D-CNNs but also promotes sustained positive learning behavior.

\section{Conclusion}
In this work, we empirically investigate the learning behavior of 1D-CNNs on TSC tasks. From accuracy degradation, we point out that deeper CNNs tend to distract the focus from LFCs and the disturbing convolution is the driving factor. To apply our findings in practice, we propose a regulatory framework to alleviate this issue. Through comprehensive experiments, we verify our findings and the effectiveness of our framework. It is worth mentioning that the goal of our work is to investigate the learning behavior of deep 1D-CNNs for TSC tasks and uncover the underlying causes of the bottleneck in these models. By sharing our findings with the community, we hope to open up new possibilities in the design of powerful TSC networks and the advancement of theoretical understanding of 1D-CNNs. 
While the verification of TCE's insights in alternative network structures (e.g., Graph Neural Networks) is still ongoing, it has exhibited great potential in exploring a variety of fascinating topics related to 1D-CNNs in the context of TSC tasks. For instance, it can assist in identifying useful frequency bands for guiding the learning of convolutional layers. Overall, it is our aspiration that the TCE can provide profound insights to the community as they move forward with their future work.

\section*{Acknowledgments}
This work was supported by the Plan Project of Ningbo Municipal Science and Technology (No. 2022S172).
\bibliographystyle{ACM-Reference-Format}
\balance
\bibliography{cikm2023}










\end{document}